\documentclass[10pt,twocolumn,letterpaper]{article}

\usepackage[pagenumbers]{cvpr} % To force page numbers, e.g. for an arXiv version

\usepackage{graphicx}
\usepackage{amsmath}
\usepackage{amssymb}
\usepackage{booktabs}
\usepackage{multirow}

\newcommand{\F}{\mathcal{F}}

\newcommand{\N}{\mathcal{N}}

\newcommand{\R}{\mathbb{R}}

\graphicspath{ {./images/} }

\usepackage[pagebackref,breaklinks,colorlinks]{hyperref}

\usepackage[capitalize]{cleveref}
\crefname{section}{Sec.}{Secs.}
\Crefname{section}{Section}{Sections}
\Crefname{table}{Table}{Tables}
\crefname{table}{Tab.}{Tabs.}

\usepackage{easy-todo}

\def\cvprPaperID{9922} % *** Enter the CVPR Paper ID here
\def\confName{CVPR}
\def\confYear{2023}

\begin{document}

\title{HashCC: Lightweight Method to Improve the Quality of the Camera-less NeRF Scene Generation}

\author{Jan Olszewski\\
University of Warsaw\\
{\tt\small janolszewski1906@gmail.com}
%\and
}

\maketitle

% \begin{figure}[b]
%     \centering
%     \subfloat[NeRF-\/-]{
%         \label{ref_label1}
%         \includegraphics[width=0.22\textwidth]{images/LLFF/NeRFmm_fern.png}
%     }
%     \subfloat[HashCC NeRF-\/-]{
%         \label{ref_label2}
%         \includegraphics[width=0.22\textwidth]{images/LLFF/ours_fern.png}
%     }
%     \caption{HashCC enables rendering more detailed images.}
%     \label{ref_label_overall}
% \end{figure}

%%%%%%%%% ABSTRACT
\begin{abstract}
    Neural Radiance Fields has become a prominent method of scene generation via view synthesis. A critical requirement for the original algorithm to learn meaningful scene representation is camera pose information for each image in a data set. Current approaches try to circumnavigate this assumption with moderate success, by learning approximate camera positions alongside learning neural representations of a scene. This requires complicated camera models, causing a long and complicated training process, or results in a lack of texture and sharp details in rendered scenes. In this work we introduce Hash Color Correction (HashCC)---a lightweight method for improving Neural Radiance Fields rendered image quality, applicable also in situations where camera positions for a given set of images are unknown.
\end{abstract}

%%%%%%%%% BODY TEXT
\section{Introduction}
\label{sec:intro}
Neural Radiance Fields~\cite{nerf} (NeRF) is a view synthesis method of generating complex 3D views from a sparse set of 2D images. The algorithm introduced in~\cite{nerf} works on a single scene and requires a collection of images with accurate camera pose estimations from which the images were taken. Then, through a training process, the information about the scene is encoded into a neural network. Using differential rendering~\cite{raytracing}, one can integrate such output along a viewing ray of interest to obtain a color value for a particular pixel in the rendered image.

The original NeRF implementation~\cite{nerf} had multiple drawbacks. The most prominent ones were (1) its speed---it takes from multiple hours to a few days to generate a single scene, and (2) images require known and precise camera positions---a feat that very few scenes had, which was a factor preventing a wide-spread usage of the algorithm. M\"uller \textit{et al.} introduced HashNeRF~\cite{hashnerf}---a novel type of encoding that greatly accelerated the training time, decreasing the scene generation time to minutes. Although the current methods produce near photorealistic results, the general application is still constrained to scenes with known camera positions. One way to remedy that is to use the COLMAP~\cite{colmap} algorithm---it is a classical computer vision algorithm capable of precomputing the camera positions. However, to obtain the precision required by NeRF or HashNeRF to produce high-quality scenes, the COLMAP algorithm needs to run for a couple of hours. 

Alternatively, in cases where camera positions are unknown or only roughly estimated~\cite{nerfmm, scnerf}, we can speak of a so-called camera-less regime. In the camera-less regime however, the vanilla NeRF architecture (used in NeRF-\/-~\cite{nerfmm}) produces blurry results. Currently, known algorithms working in the camera-less regime focus on modeling the camera in the soundest way possible~\cite{scnerf, omninerf}. Then, through a complicated training routine, gradually estimate all needed parameters of a camera model to obtain approximate camera positions with NeRF training concurrently.

Another challenge when working with NeRF models are floating artifacts and noisy color prediction. Generating a high-quality image, free of such errors, requires balancing two opposite aspects of rendering, namely: rendering sharp details and keeping the image smooth is a challenging task. To fight this issue, one usually uses some regularization techniques~\cite{regnerf, dreamfields, augnerf, dietnerf, tvloss}.

In this work, we introduce HashCC: a method to increase the expressive power of the vanilla NeRF architecture in a way that improves the quality of generated scenes and camera pose estimation in the camera-less regime. We showcase this by extending NeRF-\/- by adding a Color Correction (CC) module, consisting of a Hash Encoding layer and a Color Correction Network, and Spherical Harmonics encoding alongside Fourier encoding---modifications designed to improve the prediction quality. The HashCC method does not use any regularization, does not require modifying the training schedule, and its computational overhead is negligible. Our main focus was to develop a simple solution that can be applied to different NeRF architectures and improve the color prediction of the existing solutions. We evaluate the accuracy of HashCC on forward-facing scenes using the LLFF~\cite{llff} dataset---a setting in which camera positions are estimated from scratch. We are interested in this particular use case as estimating radiance field and camera poses simultaneously requires keeping a balance between learning general structure and the details of a scene.
%-------------------------------------------------------------------------

\subsection{Encodings}

Hash Encoding (HE) is a layer proposed in Instant Neural Graphics Primitives~\cite{hashnerf} as an alternative to Fourier Encoding (FE) layer. In contrast to FE, it has trainable parameters that are fitted together with the neural network during training. The HE layer consists of: (i) multi-resolution grids acting as multi-scale partitions of a scene, and (ii) trainable feature vectors placed in the vertices of voxels used to calculate embeddings. Voxel-level features allow producing very localized and flexible embeddings, mapping an input signal into a high-frequency domain in the exact places it is needed (e.g. regions of a scene where grass appears). This expressive strength enables a rapid training of NeRF models, which can have substantially smaller MLP layers in their architecture, due to the fact that the HE layer is also trainable.

The \emph{hash} part of the HE layer means that not every vertex gets a distinct feature vector, but rather an index which is hashed to obtain the desired feature vector. Such an approach allows for very fine partition grids of a scene, while keeping the number of required feature vectors and memory low. During training, feature vectors are optimized together with the neural network weights, which results in embeddings tailored for the network.

FE layer $\gamma$, along with a multi-layer perceptron (MLP) $\N$ with parameters $\theta_\N$, are the building blocks of the standard NeRF architecture. A non-learnable FE layer encodes the input $x$ is described by eq.~\ref{eq:fourier_embedding}

\begin{equation}
\label{eq:fourier_embedding}
\gamma(x) = (\sin(2^k x), \cos(2^k x))_{k=0}^{L-1},
\end{equation} 
where $L$ is a hyperparameter (usually in the order of 10) denoting the degree of encoding. The NeRF model $\F$, taking a point position $x$ and a viewing direction $\varphi$ as inputs, is described by eq.~\ref{eq:nerf_model}

\begin{equation}
\label{eq:nerf_model}
    \F(x, \varphi | \theta_\N) = \N(\gamma(x), \gamma(\varphi) | \theta_\N).
\end{equation} 

By default, the embedding of a viewing direction $\varphi$ uses the same FE layer used for spatial arguments $x$. However, since SH is an orthonormal basis for a space of functions defined on a sphere, it is a more natural embedding than FE. Choosing the Spherical Harmonic (SH) basis for the viewing direction embedding has already been adopted in the newest research~\cite{refnerf, plenoctrees, plenoxels ,hashnerf} trying to improve on the original NeRF algorithm. We report improvements in rendered images quality and camera pose estimation with respect to the usage of FE and SH. 

\subsection{Camera pose estimation and camera-less regime}
\label{sec:intro:cameraless}
In real life scenarios we rarely have the ground truth information about the position of camera when a picture was taken. The common name for algorithms addressing this problem is Structure from Motion (SfM). One of the most commonly used SfM algorithm is COLMAP~\cite{colmap}. COLMAP, given a set of images and enough time, estimates camera positions accurately and its predictions are treated as ground truth for training NeRF models. In camera-less regime, we want to avoid this step by jointly learning camera poses and radiance fields.

Camera positions can be modeled in multiple ways. One of the simplest camera models is a so-called pinhole camera model. It consists of a camera pose and a focal length information. The pose is represented as an $SE(3)$ matrix $[R\mid t]$, where $R\in SO(3)$ and $t\in \R^3$ describes rotation and translation of the camera in the world coordinates, respectively. The focal length is described by a pair of scalars $f_x, f_y$ describing how strongly a lens bends light in corresponding $x$ and $y$ dimensions. Such a simple model provides decent results and is very easy to train. Unfortunately, real lenses cannot be properly described by a simple pinhole model. The cameras we use day-to-day introduce many non-linear distortions, discussed extensively in SCNeRF~\cite{scnerf}. Moreover, the images can come from radically different lenses, like fish-eye lens, which was addressed in OmniNeRF~\cite{omninerf}. Finally, we could also take into account the fact that pixels in images are not infinitesimally small---a source of light projected onto a pixel should not be modelled as a ray, but rather a frustum, which was discussed in Mip-NeRF~\cite{mipnerf}. 

NeRF-\/-~\cite{nerfmm} and BaRF~\cite{barf} were the first attempts to develop camera-less NeRF. Both works address the problem by using a minimal possible setup, namely: (i) an MLP for the neural representation of a scene and (ii) a simple pinhole camera model. NeRF-\/- showed that one can jointly optimize camera model and radiance fields MLP, by simply adding camera as a trainable module. Nevertheless, resulting scenes are of substantially lower quality than those presented in original NeRF paper, where camera poses are a known prior. BaRF showed that if we set the training schedule more carefully, employing coarse to fine optimization, we can obtain very good camera pose estimation. However, BaRF assumes a known camera focal information and works on smaller resolution images ($480\times640$).

\section{Methods}

% \todo{MS: I'd call it "Methods"}
% \todo{PR: Extra parameterization tricks should be moved to this section, which should be bigger and provide all the necessary information to reproduce our results as well as describing the trianing setup}

% \subsection{HashCC overview}
Our approach to address the camera-less training problem is to use the simplest setup---introduce a trainable pinhole camera model and to use an MLP neural network to estimate radiance fields. On top of this architecture, we add a small neural network, which utilizes a HE layer for correcting the color outputs of the main model. Our method can be easily extended to exploit different camera models and architectures.

\subsection{Implementation details}
\begin{figure*}[t]
        \centering
        \includegraphics[width=\textwidth]{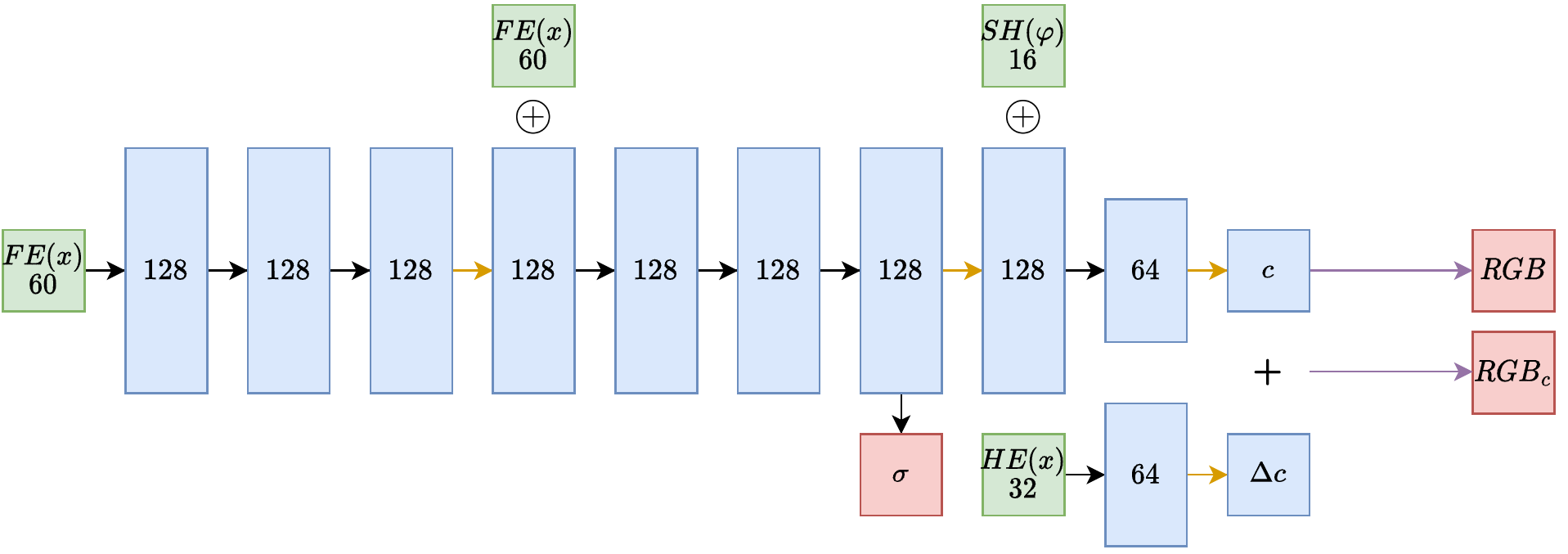}
        \caption{ The NeRF implementation used in our paper with the color correction module below. Green: encoded input. Blue: MLP hidden layers, with channel dimension shown inside. Red: radiance and density outputs. $\oplus$ denotes feature concatenation, $+$ denotes usual addition. The solid black arrow denotes layers with ReLU activation. The purple arrow
denotes applying sigmoid activation. The orange arrow denotes layers without activation. The figure style is borrowed from the NeRF and NeRF-\/- papers so readers can easily make comparisons.}
        \label{detailed_architecture}
\end{figure*}

Our implementation of the main model builds on the NeRF-\/- architecture (Fig.~\ref{detailed_architecture}). Similarly, we use 10-degree FE for position input $x$, resulting in 60-element embedding (three coordinates and $\sin, \cos$ function per each degree). The embedded input is passed to a nine-layer MLP neural network, whose first eight hidden layers have 128 nodes, and the last one has 64 nodes in the main network. Biases of the density prediction layer are initialized to be 0.1 and the main color prediction layer to be 0.02, following a common practice.

Each camera is represented with a pinhole model initialized at the origin, looking in $-z$ direction with focal length initialized as width and height of the image. Naively optimizing focal parameters results in numerical issues, as both width and height of an image are of much larger magnitude than other learnable parameters. To fight this issue we learn the focal expressed as a scaling of width and height dimensions instead. Following the recipe of NeRF-\/-, we use second order parameterization of the scaling parameter to improve the results. Differently to NeRF-\/-, we encode viewing direction $\varphi$ using SH of degree 4 (resulting in 16-element embedding, since SH polynomial of degree $k$ has $k^2$ coefficients).

To extend the NeRF-\/- model, we include the Color Correction Network. The first component is the HE layer with 16 resolution levels and two-dimensional feature vectors attached to vertices of partition grids with a maximum of 512 feature vectors at each resolution level. The second component is a two-layer MLP neural network with one hidden layer of width 64. 

Additionally, we use Normalized Device Coordinates (NDC)~\cite{nerf}, which is a standard method used for forward-facing scenes. The NDC parameterization transforms unbounded scene dimensions into a $[-1,1]^3$ cube in the hyperbolic fashion. Importantly, it squashes the space distant from the camera into a smaller chunk of the cube and stretches out a nearby region onto a larger part of the cube. Sampling points uniformly along a direction vector in NDC coordinates results in sampling more points in the foreground of the image and fewer in the background.

HashCC was implemented in PyTorch~\cite{pytorch}, based on the official NeRF-\/- implementation\footnote{\href{https://github.com/ActiveVisionLab/nerfmm}{https://github.com/ActiveVisionLab/nerfmm}}.

\subsection{Dataset}
We conduct experiments on the LLFF dataset~\cite{llff} which consists of eight forward-facing scenes each containing between 20 and 62 images. In all experiments, we follow the official pre-processing procedures and train/test splits, i.e., the resolution of the training images is $756\times1008$, and every eighth image is used as a test image.

\subsection{Training}
A standard practice when training NeRFs is to perturb density prediction for regularization purposes, we do not employ such strategy. Similarly to NeRF-\/-, we do not practice hierarchical sampling either. Unlike BaRF~\cite{barf}, we train the whole architecture simultaneously---without special learning rate schedulers, simplifying the whole process.

\begin{figure*}[t]
        \centering
        \includegraphics[width=\textwidth]{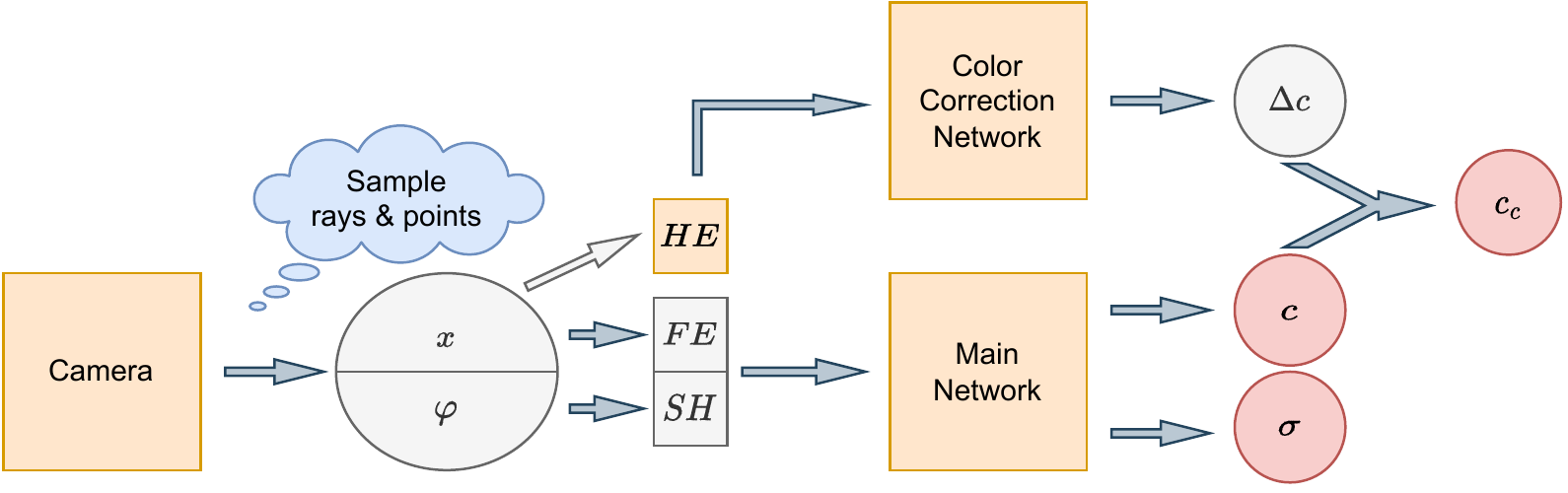}
        \caption{Visualization of a forward pass through a model working in the camera-less regime with color correction. Yellow boxes are the trainable modules. Red circles are the outputs to be integrated with the differential rendering formula. Blue-colored arrows denote operations that are included in the backward pass. $HE$ - Hash Encoding layer, $FE$ - Fourier Encoding layer, $SH$ - Spherical Harmonics encoding layer.}
        \label{HCC_forward}
\end{figure*}

First, we sample a collection of points $x$ with a viewing direction $\varphi$ from a camera module (denoted as ``Camera'' in Fig.~\ref{HCC_forward}). Then, $x$ and $\varphi$ are passed to two neural networks: the Main Network and the Color Correction Network. The Main Network is a typical NeRF architecture, using FE and SH layers to process the input, and returning color $c$ value and density $\sigma$ of points sampled along the ray for each pixel in the rendered image. The second network, called Color Correction Network, uses HE and returns color correction to the main output $\Delta c$. Final color prediction is obtained by a simple addition of the outputs of two networks: $c_c = c + \Delta c$ and the final output $(c_c, \sigma)$ is returned. The final output can be then used to compute the pixel color prediction $\tilde{p_c}$ using the differential rendering formula~\cite{raytracing}. During training, both outputs $(c, \sigma)$ and $(c_c, \sigma)$ are used to compute their corresponding predictions $\tilde{p}$ and $\tilde{p_c}$, respectively.

% Let $C(\cdot \mid \theta_C)$ be a trainable camera module, from which we sample a collection of points $x$ with a viewing direction $\varphi$.
% We do a forward pass through the main model resulting in the main output 

% \begin{equation}
%     (c, \sigma) = \F(x, \varphi \mid \theta_\N).
% \end{equation} Let's define our color correction network \begin{equation}
%     \Ha(x \mid \theta_\Ha) = \N'( H( x \mid \theta_H) \mid \theta_{\N'})
% \end{equation} for some shallow neural network $\N'$ and Hash Encoding layer $H$. A forward pass through $\Ha$ results in a color correction to the main output
% \begin{equation}
%     \Delta c = \Ha(x \mid \theta_\Ha).
% \end{equation}

%Moreover, in training iteration we sample 1024 random rays from a concrete image and 128 points along each ray.
To construct a training batch we first sample a single image from the training set and random grid of 1024 rays (32 random rows and 32 random columns). Then, we sample 128 random points along each ray, resulting in a $1\times1024\times128$ input tensor. The training is performed for 10000 epochs (resulting in [$10000\times$number of images in training set] number of iterations) using Adam optimizer~\cite{adam} with initial learning rate (LR) of $10^{-3}$. The Main Network and Color Correction Network with the HE layer are trained with exponential LR decay, decreasing the LR to $10^{-5}$ towards the end of training. For the camera model, we use the same optimizer setup but with the decay parameter set to reduce the LR to $10^{-8}$ towards the end of training.

During training, we minimize the combined loss function $\|p-\tilde{p}\|^2 + \|p - \tilde{p_c}\|^2$ ensuring that the Main Model learns a rough structure of a scene and the Color Correction Network fills up the details. In order to improve the stability of training, we backpropagate gradients to the camera model only from the main prediction $\tilde{p}$. This means that in the forward pass of the HE layer of the Color Correction Network we detach input points from the computational graph, learning only to correct predictions of the main model instead of steering the camera. We empirically found that gradients of the HE layer with respect to input are highly non-regular, making it difficult to learn the camera positions.

\subsection{Evaluation}
\label{sec:methods:evaluation}
The model was evaluated following the scheme introduced in NeRF-\/-. First, we compute the $SIM(3)$ transformation between estimated and ground truth (estimated by COLMAP) set of camera positions using the ATE Toolbox~\cite{ate}. The $SIM(3)$ transformation consists of (i) translation vector, (ii) rotation matrix, (iii) scale number. Then, we initialize cameras to be used for evaluation of novel view synthesis with test poses transformed by the similarity map. We need to do this because learned camera poses lie in a different coordinate system than test poses. To measure the quality of images rendered from novel views and exclude error factors induced by inaccurate camera pose and $SIM(3)$ map estimation, we perform additional gradient descent optimization of the camera parameters with the rest of model parameters being frozen. Such a corrected camera view is then used to render an evaluation image to be compared with a test image.

To assess the prediction quality of our method we compare it against NeRF-\/-. Since our method builds on NeRF-\/-, such comparison allows us to directly assess the influence of the improvements we introduced in the HashCC method, which is the main contribution of this work.

We evaluate the quality of rendered images, but also correctness of estimated camera poses.
Image quality is assessed with 3 metrics, that capture different aspects of rendering:
\begin{itemize}
    \item Peak Signal-to-Noise Ratio (PSNR)
    \item Structural Similarity Index Measure (SSIM)
    \item Learned Perceptual Image Patch Similarity (LPIPS)
\end{itemize}

PSNR~\cite{ssim_vs_psnr} is a metric based on pixel-wise MSE, taking into account maximal value of used RGB representation and normalizing the score accordingly. SSIM~\cite{ssim} is less sensitive to small translations of the rendered objects---it puts more emphasis on higher level features like: luminance, contrast, and shape visible in the picture. Finally, LPIPS~\cite{lpips} compares pre-activations of a classifier neural network (usually VGG or AlexNet trained on ImageNet) applied to grand truth and generated images. This metric expresses semantic agreement between the two images, contrary to previous metrics that focus on a pixel-level information or low-level features. In this work we use VGG to calculate LPIPS for easier comparison with NeRF-\/-, which also uses this network.

To evaluate the accuracy of the camera pose estimation, we compare the predictions of the model against those created by COLMAP~\cite{colmap}. Using the ATE toolbox~\cite{ate} we compute:
\begin{itemize}
    \item $\Delta$ R$^{\circ}$ -- rotation difference (in degrees)
    \item $\Delta$ T -- translation difference
\end{itemize} 

The NeRF-\/- scores were obtained by running evaluation scripts and using model checkpoints provided by the authors in their repository\footnote{\href{https://github.com/ActiveVisionLab/nerfmm}{github.com/ActiveVisionLab/nerfmm}}. We have used NeRF-\/- models in which camera focal was parameterized by a pair of independent scalars, as the same setup was used in our method.

\section{Results}
\label{sec:results}

In this section, we provide the results of our experiments. We evaluate our method on the LLFF~\cite{llff} dataset, comparing against the NeRF-\/- baseline.

For six out of eight scenes, our method provides the same or better image quality than the vanilla NeRF-\/- according to all three metrics (Table~\ref{tab:results}). For the two remaining scenes, ``T-rex'' and ``Room'' our method achieves a better LPIPS score, but worse PSNR and SSIM than NeRF-\/-. Qualitatively, the difference between NeRF-\/- and HashCC on two scenes, ``Ferns'' and ``T-rex'', is shown in Fig.~\ref{fig:fern_trex}. Despite the fact that for the PSNR, SSIM, $\Delta$R$^{\circ}$, and $\Delta$T our method yielded lower score, the generated images are sharper, exhibit more detail and have more pronounced texture, which is especially visible on the pelvis of the T-rex. The qualitative comparison of the remaining scenes can be found in the Supplementary Materials.

Our method also outperforms NeRF-\/- in terms of camera pose estimation (see the right side of Table~\ref{tab:results}). For six out of eight scenes, the rotation angle predicted by our method is closer to the reference one than the NeRF-\/- estimate (smaller $\Delta$ R$^{\circ}$). Also, the same scenes, the translation difference ($\Delta$ T) for our method is the same or smaller than for vanilla NeRF-\/-.

\begin{table*}[t]
  \centering
\begin{tabular}{c | cc c cc c cc | cc c cc}
\toprule
\multirow{3}{*}{Scene} & \multicolumn{8}{c}{View synthesis quality} & \multicolumn{5}{|c} {Camera pose registration}  \\
&
\multicolumn{2}{c}{PSNR $\uparrow$} &
&
\multicolumn{2}{c}{SSIM $\uparrow$} &
&
\multicolumn{2}{c}{LPIPS $\downarrow$} &
\multicolumn{2}{|c}{$\Delta$ R$^{\circ}$ $\downarrow$} &
&
\multicolumn{2}{c}{$\Delta$ T $\downarrow$} \\
&
NeRF-\/- & ours
&&
NeRF-\/- & ours
&&
NeRF-\/- & ours
& 
NeRF-\/- & ours
&& 
NeRF-\/- & ours\\
\midrule
Fern & 21.78&\textbf{22.27} && 0.62&\textbf{0.66} && 0.50&\textbf{0.40} & 1.43&\textbf{1.38} && 0.008&\textbf{0.007}\\
Flower & 25.25&\textbf{25.89} && 0.71&\textbf{0.75} && 0.37&\textbf{0.29} & \textbf{4.43}&4.89 && 0.012&0.012\\
Fortress & 26.25&\textbf{26.85} && 0.65&\textbf{0.71} && 0.46&\textbf{0.33} & 2.63&\textbf{2.16} && 0.062&\textbf{0.044}\\
Horns & 23.08&\textbf{23.53} && 0.63&\textbf{0.66} && 0.50&\textbf{0.42} & 9.08&\textbf{3.41} && 0.056&\textbf{0.016}\\
Leaves & 18.80&\textbf{19.93} && 0.52&\textbf{0.64} && 0.47&\textbf{0.37} & 6.89&\textbf{3.16} && 0.006&\textbf{0.004}\\
Orchids & 16.48&\textbf{16.50} && 0.37&0.37 && 0.56&\textbf{0.55} & 4.73&\textbf{4.69} && 0.019&\textbf{0.018}\\
Room & \textbf{25.75}&25.64 && 0.83&0.83 && 0.44&\textbf{0.40} & 3.10&\textbf{3.09} && \textbf{0.013}&0.023\\
T-rex & \textbf{22.55}&22.37 && \textbf{0.72}&0.71 && 0.44&\textbf{0.41} & \textbf{6.15}&6.67 && \textbf{0.013}&0.016\\
\bottomrule
\end{tabular}

  \caption{Quantitative comparison between NeRF-\/- and our method. First three metric columns (PSNR, SSIM, LPIPS) describe rendered image quality in comparison to ground truth image. Two following metric columns ($\Delta$ R$^{\circ}$, $\Delta$ T) describe the camera pose estimation error in comparison to the COLMAP estimation. For each scene and metric we \textbf{bold} the better score of the two compared models.}
  \label{tab:results}
\end{table*}

We have also performed an ablation study to measure the effect of improving the NeRF-\/- method. We have modified NeRF-\/- to apply the SH encoding to the directional inputs $\varphi$, without adding the Color Correction network. Our results show that the SH basis, being a natural embedding for unit vectors describing directions $\varphi$ improves camera pose estimation, especially the rotational component as can be seen in Table~\ref{tab_SH}. At the same time, after adding the Color Correction network, the model seems to achieve worse scores on the camera pose estimation. With the Color Correction network, the model has an easier way to improve the loss function than to optimize the camera position, so it focuses on generating the colors properly instead of getting the position right, slightly lowering the $\Delta$R$^{\circ}$, and $\Delta$T scores. 

\begin{table*}[t]
  \centering

\begin{tabular}{c | ccc c ccc | ccc c ccc}
\toprule
\multirow{3}{*}{Scene} & \multicolumn{7}{c}{View synthesis quality} & \multicolumn{7}{|c} {Camera pose registration}  \\
&
\multicolumn{3}{c}{PSNR $\uparrow$} &
&
\multicolumn{3}{c}{LPIPS $\downarrow$} &
\multicolumn{3}{|c}{$\Delta$ R$^{\circ}$ $\downarrow$} &
&
\multicolumn{3}{c}{$\Delta$ T $\downarrow$} \\
&
FE & SH & ours
&&
FE & SH & ours
& 
FE & SH & ours
&&
FE & SH & ours\\

\midrule
Fern & 21.78 & 21.80 & \textbf{22.27} && 0.50 & \textbf{0.40} & \textbf{0.40} & 1.43 & \textbf{1.25} & 1.38 && 0.008 & \textbf{0.007} & \textbf{0.007} \\

Fortress & 26.25 & \textbf{26.85} & \textbf{26.85} && 0.46 & 0.41 & \textbf{0.33} & 2.63 &\textbf{2.01} & 2.16 && 0.062 & \textbf{0.036} & 0.044 \\

Leaves & 18.80 & 18.83 & \textbf{19.93} && 0.47 & 0.48 & \textbf{0.37} & 6.89 &\textbf{2.60} & 3.16 && 0.006 &\textbf{0.003} & 0.004 \\

Room & \textbf{25.75} & 25.71 & 25.64 && 0.44 & 0.44 & \textbf{0.40} & 3.10 & \textbf{3.08} & 3.09 && \textbf{0.013} & 0.016 & 0.023 \\

T-rex & 22.55 & \textbf{22.58} & 22.37 && 0.44 & 0.45 & \textbf{0.41} & 6.15 & \textbf{5.35} & 6.67 && 0.013 & \textbf{0.012} & 0.016 \\
\bottomrule

\end{tabular}

%Fern & 21.80 & 0.62 & 0.49 & 1.25 & 0.007\\
%Flower & 21.39 & 0.71 & 0.37 & 4.00 & 0.010\\
%Fortress & 26.86 & 0.69 & 0.41 & 2.01 & 0.036\\
%Horns* & NaN & NaN& NaN& NaN& NaN\\
%Leaves & 18.83 & 0.52 & 0.48 & 2.60& 0.003\\
%Orchids & 16.52 & 0.37 & 0.57 & 4.89 & 0.019\\
%Room & 25.71 & 0.83 & 0.44 & 3.08 & 0.016\\
%T-rex & 22.58 & 0.72 & 0.45 & 5.35 & 0.012\\
  \caption{Quantitative results of evaluating NeRF-\/- architecture with usual Fourier Encoding of directional inputs (FE columns) against NeRF-\/- with Spherical Harmonics (SH columns), and our method (NeRF-\/- with SH and the Color Correction network). First two metric columns (PSNR, LPIPS) describe rendered image quality in comparison to ground truth image. Two following metric columns ($\Delta$ R$^{\circ}$, $\Delta$ T) describe camera pose estimation error in comparison to COLMAP estimation. For each scene and metric we \textbf{bold} the better score of the three compared models. \\
  %* - Training diverged
  }
  \label{tab_SH}
\end{table*}

% \begin{figure*}[!h]
% \centering
% \captionsetup[subfigure]{labelformat=empty}
%     \centering
%     \subfloat{
%         \label{ref_label1}
%         \includegraphics[width=0.5\textwidth]{images/LLFF/shnerfmm_fern.jpg}
%     }
%     \subfloat{
%         \label{ref_label2}
%         \includegraphics[width=0.5\textwidth]{images/LLFF/shnerfmm_fortress.jpg}
%     }
    
%     \subfloat[Nerf-\/-]{
%         \label{ref_label1}
%         \includegraphics[width=0.5\textwidth]{images/LLFF/shnerfmm_leaves.jpg}
%     }
%     \subfloat[HashCC NeRF-\/-]{
%         \label{ref_label2}
%         \includegraphics[width=0.5\textwidth]{images/LLFF/shnerfmm_room.jpg}
%     }
    
%     \caption{Images rendered with NeRF-\/- where SH encoding was used for embedding direction $\varphi$ instead of usual FE. See Section~\ref{sec:discussion} for detailed discussion about the results and Table~\ref{tab_SH} for quantitative comparison against original NeRF-\/-.}
%     \label{fig:shnerfmm}
% \end{figure*}

\section{Discussion}
\label{sec:discussion}

\begin{figure*}[ht]
\centering
\captionsetup[subfigure]{labelformat=empty}
    \centering
    \subfloat[NeRF-\/-, Fern]{
        \label{ref_label1}
        \includegraphics[width=0.49\textwidth]{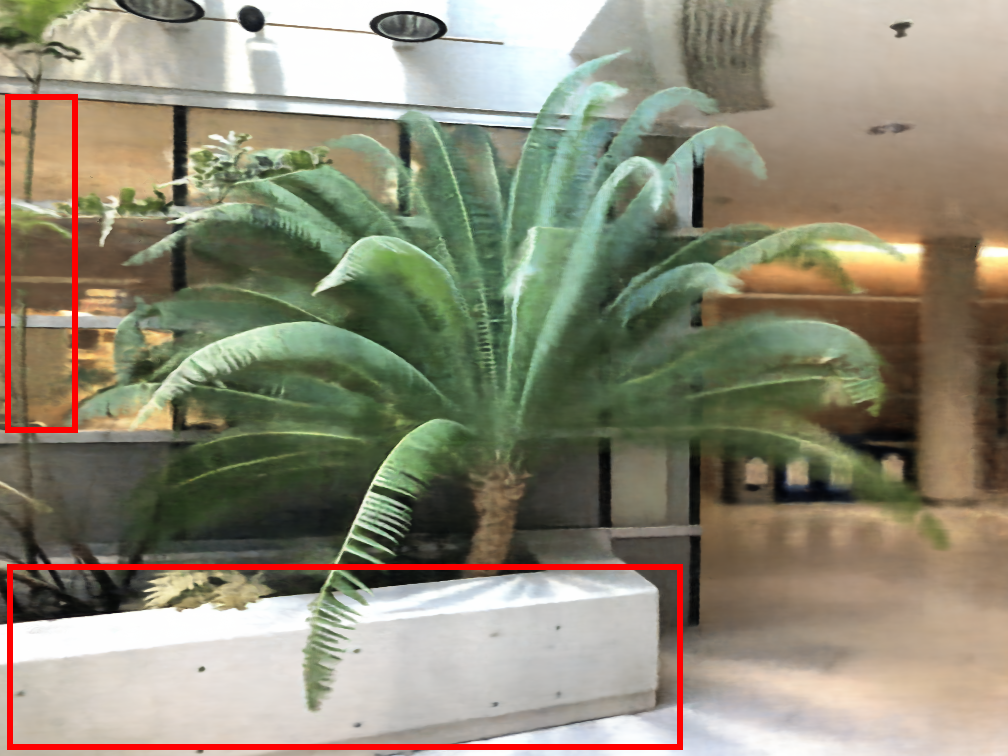}
    }
    \subfloat[NeRF-\/-, T-rex]{
        \label{ref_label2}
        \includegraphics[width=0.49\textwidth]{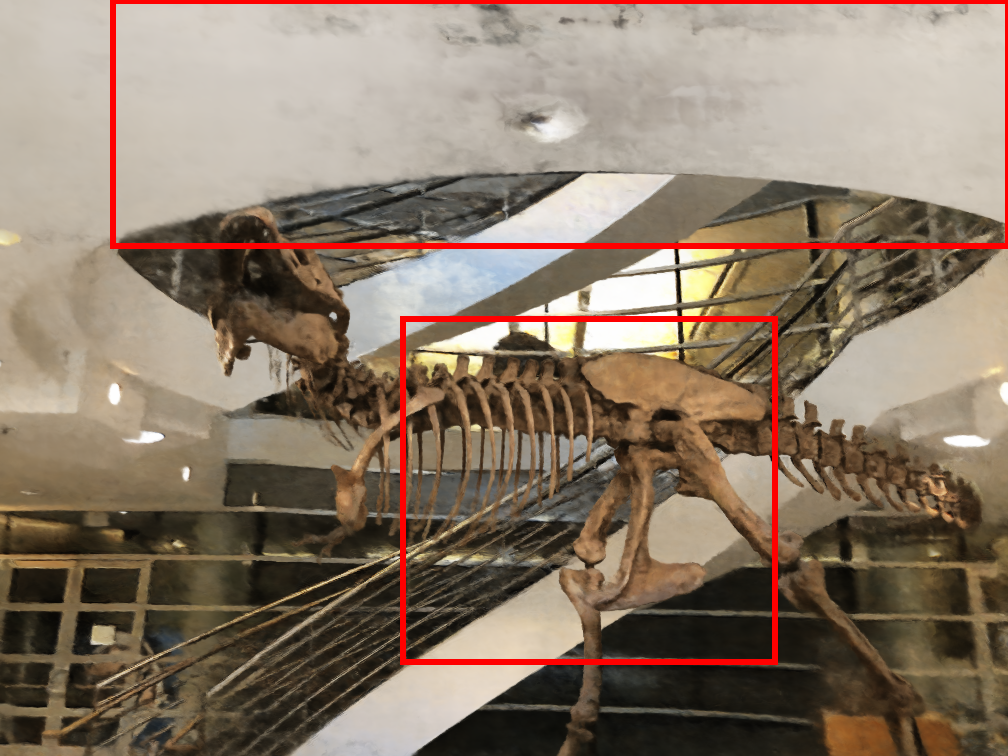}
    }
    
    \subfloat[NeRF-\/- + SH, Fern]{
        \label{ref_label1}
        \includegraphics[width=0.49\textwidth]{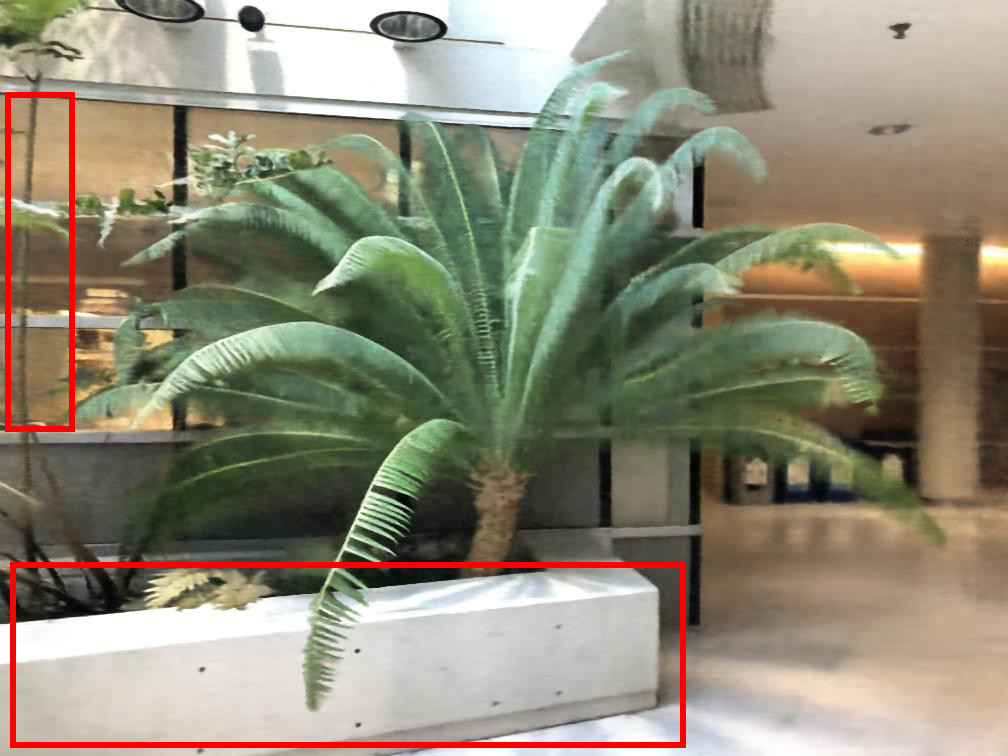}
    }
    \subfloat[NeRF-\/- + SH, T-rex]{
        \label{ref_label2}
        \includegraphics[width=0.49\textwidth]{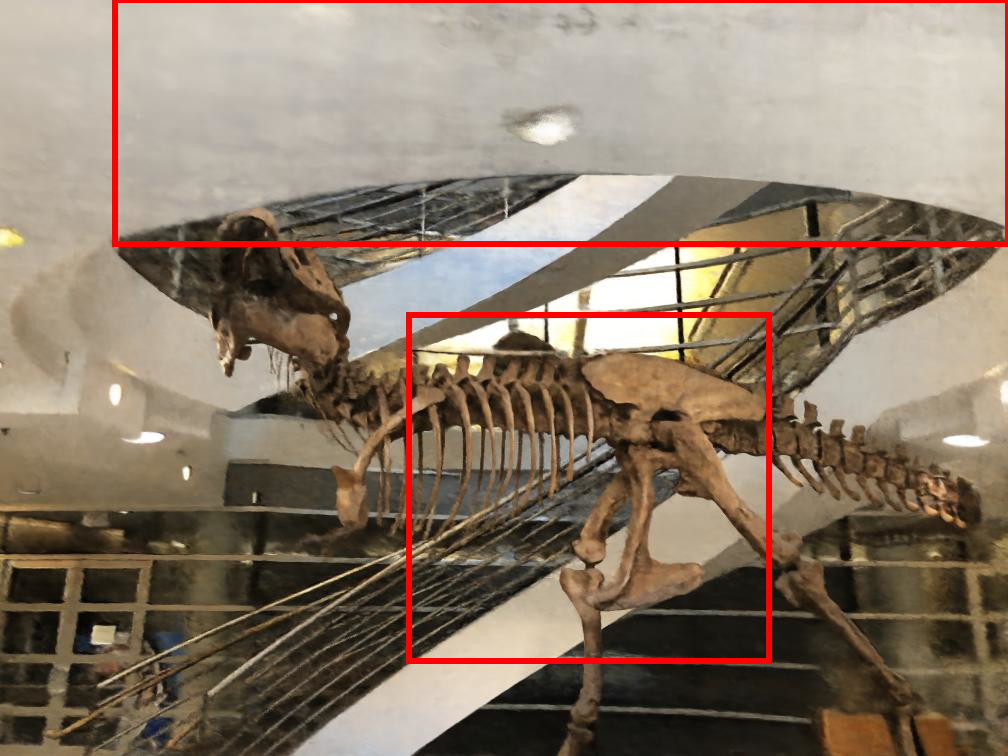}
    }
    
    \subfloat[ours, Fern]{
        \label{ref_label1}
        \includegraphics[width=0.49\textwidth]{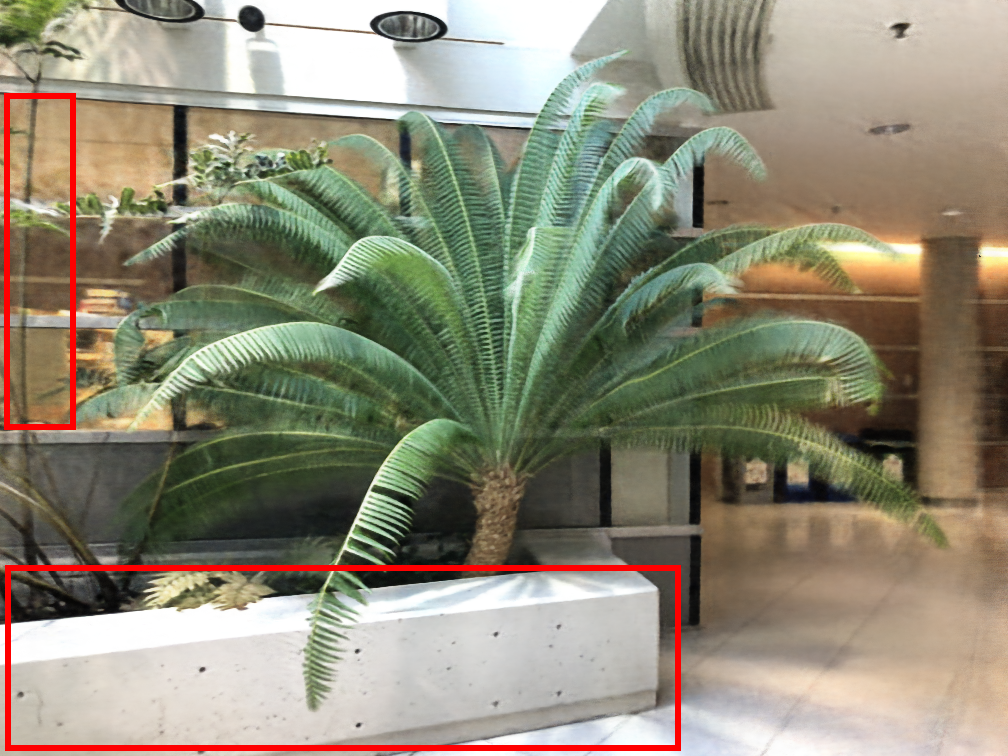}
    }
    \subfloat[ours, T-rex]{
        \label{ref_label2}
        \includegraphics[width=0.49\textwidth]{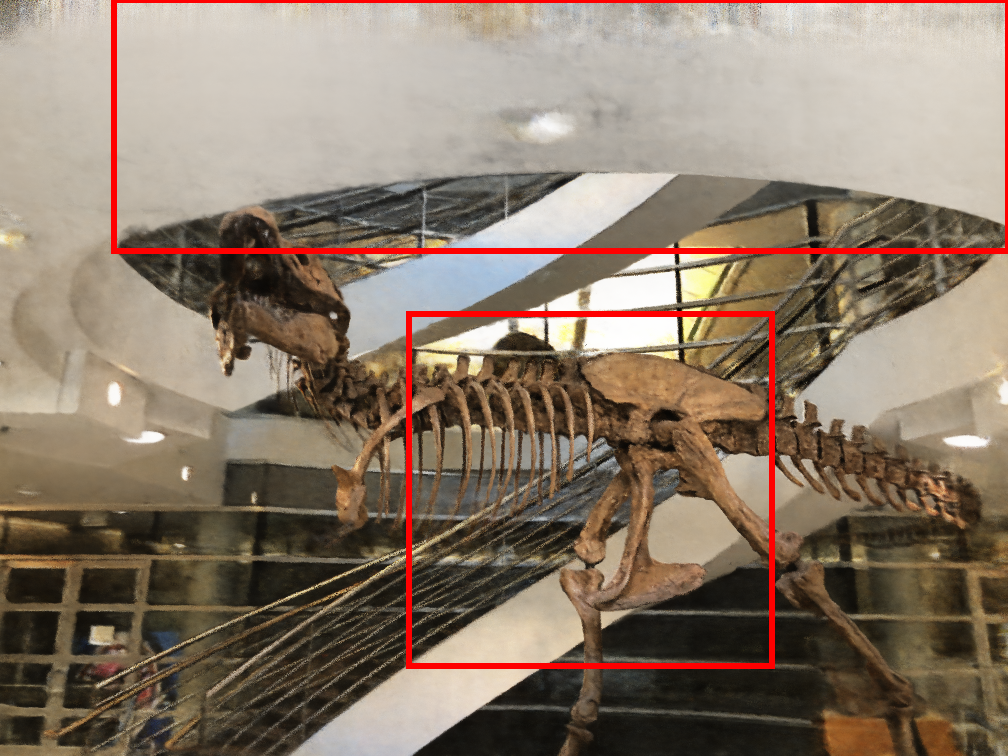}
    }
    
    \caption{Qualitative comparison of novel view synthesis between vanilla NeRF-\/- algorithm, NeRF-\/- with Spherical Harmonics used as embedding of directional inputs $\varphi$ instead of Fourier embedding and our version with the color correction (HashCC NeRF-\/-) on the ``Fern'' (left) and ``T-rex'' (right) scenes.}
    \label{fig:fern_trex}
\end{figure*}

Our method improves the quality of rendered images and provides improvement on camera pose estimates concerning the main model in the majority of considered scenes. The improvement is especially noticeable in the ``Fern'' scene (Fig.~\ref{fig:fern_trex}) where our method compared to NeRF-\/- is successfully able to separate distinct leaves of the fern. Less noticeable things are: NeRF-\/- was unable to render part of the branch near the left edge of the image, while our method kept more detail (black dots) on the wall below the fern. This is also visible in the quantitative comparison (Table~\ref{tab:results}), namely the LPIPS score enjoys big improvement with the support of significant improvement of PSNR and SSIM metrics. Considering the ``T-rex'' scene (Fig.~\ref{fig:fern_trex}): the PSNR and SSIM metrics (Table~\ref{tab:results}) suggest that our method performs worse than the baseline. If we focus on upper edge of the image we can localize the area which might be responsible for the poor PSNR and SSIM scoring of our method, which produces noisy results in the areas of scene unseen in the training data. This is expected, since HE used in Color Correcting Network is highly local function and, by its nature, cannot extrapolate to unseen areas of the scene. However, considering the upper part of the image, our method produces clearer white ceiling. Moreover, the upper barrier in the image produced by baseline NeRF-\/- is missing some ribs, but our method manages to render them correctly. If we focus on the bones of the skeleton, especially the pelvis, we see that the usage of HashCC results in crispier, more detailed texture, which is reflected in the higher LPIPS score (Table~\ref{tab:results}).

%Other remarkable result is ``Horns'' scene. The usage of HashCC results in crispier, more detailed texture of the skull in the center of the image and also the ones on the sides. Another noticeable thing is the floor texture, which was smoothed-out by NeRF-\/- algorithm, where our method tries to keep the original texture of the carpet intact. Again, in Table~\ref{tab:results} we see an increase in all view synthesis quality metrics, the most significant improvement being the LPIPS score.

%For some scenes, our method inherently suffers from the usage of underlying NeRF-\/- algorithm. Although our method improves view synthesis quality of rendered images, it does so only with respect to the details. If we consider the ``Orchids'' scene (Fig.~\ref{fig:orchids}) we can see that our method only improves the quality of rendered petals, leaving surroundings of the orchids with nearly the same poor quality as the ones rendered by NeRF-\/-.

We present our results in the context of camera-less NeRF, building our solution on top of NeRF-\/-. However, we expect that HashCC can be also applied to other scenarios, e.g. Light Field Neural Rendering~\cite{lightfields}, DreamFields~\cite{dreamfields}, RawNeRF~\cite{rawnerf} and other architectures, where introducing HE as a layer in the main model is not so straightforward, e.g. Mip-NeRF~\cite{mipnerf}, Ref-NeRF~\cite{refnerf}.

Table~\ref{tab_SH} contains the view synthesis quality and camera pose registration metrics of the original NeRF-\/- architecture with FE embedding for directional inputs, NeRF-\/- with SH embeddings for directional inputs $\varphi$, and our method on the selected scenes from Table~\ref{tab:results}. Although we use the same camera model as NeRF-\/-, adding SH encoding largely improved the camera pose estimation, which underpins the importance of selecting appropriate embeddings for the input data. The view synthesis quality metrics in the SH case also seem to improve slightly. Most noticeably, LPIPS score on ``Fern'' and ``Fortress'' scenes is substantially better, and the ``Fortress'' scene improving on PSNR too. However, the main quality improvement is gained by adding the Color Correction network---which represents the full HashCC method. A comparison of the rendered images on Fig.~\ref{fig:fern_trex} yields that the images produced by NeRF-\/- with SH embedding also suffer from lack of detail and texture, ensuring that the Color Correction part of our method is responsible for the visible rendered image quality improvement, produced by HashCC NeRF-\/- architecture.

Even though the improvement in camera pose estimation is noticeable, it is only incremental. Our solution still suffers from limitations of the underlying camera model. Learning camera poses of a scene captured in the non-forward-facing regime remains a challenge and an exciting potential direction of future research. Current solutions working in camera-less regime circumnavigate this obstacle by only roughly estimating camera poses with the quick usage of COLMAP and then denoising obtained camera pose initializations in the training process. Using a more sophisticated camera model, injecting prior knowledge about the shape of camera trajectory or abusing the continuity of the camera trajectory might help to solve this problem.

\section{Conclusion}

In this work we introduce HashCC: a computationally lightweight color correction method one can apply to improve the quality of NeRF rendered images. We focus on camera-less regime where currently used methods suffer either from over-smoothing or employ complicated training schedules and camera models, resulting in long training times. HashCC benefits from a color correction module, that employs HE layer and a shallow neural network, that predicts a correction term to the output of the main neural network.

We show that HashCC can be used in challenging scenarios like the camera-less regime, which often suffers from an unstable training process and overfitting. We showcase this by extending NeRF-\/- with our HashCC module and evaluating our solution on the LLFF dataset. Our results prove that HashCC adds value to the vanilla NeRF-\/- model, improving both rendered image quality and camera pose estimation for the majority of considered scenes. At the same time, our method is generic enough to be easily exploitable by other NeRF implementations.

\clearpage
%%%%%%%%% REFERENCES
{\small
\bibliographystyle{ieee_fullname}
\bibliography{egbib}

\begin{thebibliography}{10}\itemsep=-1pt

\bibitem{mipnerf}
Jonathan~T. Barron, Ben Mildenhall, Matthew Tancik, Peter Hedman, Ricardo
  Martin-Brualla, and Pratul~P. Srinivasan.
\newblock Mip-nerf: A multiscale representation for anti-aliasing neural
  radiance fields, 2021.

\bibitem{augnerf}
Tianlong Chen, Peihao Wang, Zhiwen Fan, and Zhangyang Wang.
\newblock Aug-nerf: Training stronger neural radiance fields with triple-level
  physically-grounded augmentations.
\newblock In {\em CVPR}, 2022.

\bibitem{ssim_vs_psnr}
Alain Horé and Djemel Ziou.
\newblock Image quality metrics: Psnr vs. ssim.
\newblock In {\em 2010 20th International Conference on Pattern Recognition},
  pages 2366--2369, 2010.

\bibitem{dreamfields}
Ajay Jain, Ben Mildenhall, Jonathan~T. Barron, Pieter Abbeel, and Ben Poole.
\newblock Zero-shot text-guided object generation with dream fields.
\newblock 2022.

\bibitem{dietnerf}
Ajay Jain, Matthew Tancik, and Pieter Abbeel.
\newblock Putting nerf on a diet: Semantically consistent few-shot view
  synthesis.
\newblock In {\em ICCV}, pages 5885--5894, October 2021.

\bibitem{omninerf}
Sebastian Knorr Christine~Guillemot Kai~Gu, Thomas~Maugey.
\newblock Omni-nerf: Neural radiance field from 360° image captures.
\newblock {\em ICME}, 2022.

\bibitem{raytracing}
James Kajiya and Brian von herzen.
\newblock Ray tracing volume densities.
\newblock {\em ACM SIGGRAPH Computer Graphics}, 18:165--174, 07 1984.

\bibitem{adam}
Diederik~P. Kingma and Jimmy Ba.
\newblock Adam: A method for stochastic optimization, 2014.
\newblock cite arxiv:1412.6980Comment: Published as a conference paper at the
  3rd International Conference for Learning Representations, San Diego, 2015.

\bibitem{barf}
Chen-Hsuan Lin, Wei-Chiu Ma, Antonio Torralba, and Simon Lucey.
\newblock Barf: Bundle-adjusting neural radiance fields.
\newblock In {\em ICCV}, 2021.

\bibitem{rawnerf}
Ben Mildenhall, Peter Hedman, Ricardo Martin-Brualla, Pratul~P. Srinivasan, and
  Jonathan~T. Barron.
\newblock {NeRF} in the dark: High dynamic range view synthesis from noisy raw
  images.
\newblock {\em arXiv}, 2021.

\bibitem{llff}
Ben Mildenhall, Pratul~P. Srinivasan, Rodrigo Ortiz-Cayon, Nima~Khademi
  Kalantari, Ravi Ramamoorthi, Ren Ng, and Abhishek Kar.
\newblock Local light field fusion: Practical view synthesis with prescriptive
  sampling guidelines.
\newblock {\em TOG}, 38(4), jul 2019.

\bibitem{nerf}
Ben Mildenhall, Pratul~P. Srinivasan, Matthew Tancik, Jonathan~T. Barron, Ravi
  Ramamoorthi, and Ren Ng.
\newblock Nerf: Representing scenes as neural radiance fields for view
  synthesis.
\newblock In {\em ECCV}, 2020.

\bibitem{hashnerf}
Thomas M\"uller, Alex Evans, Christoph Schied, and Alexander Keller.
\newblock Instant neural graphics primitives with a multiresolution hash
  encoding.
\newblock {\em TOG}, 41(4):102:1--102:15, July 2022.

\bibitem{regnerf}
Michael Niemeyer, Jonathan~T. Barron, Ben Mildenhall, Mehdi S.~M. Sajjadi,
  Andreas Geiger, and Noha Radwan.
\newblock Regnerf: Regularizing neural radiance fields for view synthesis from
  sparse inputs.
\newblock In {\em CVPR}, 2022.

\bibitem{pytorch}
Adam Paszke, Sam Gross, Francisco Massa, Adam Lerer, James Bradbury, Gregory
  Chanan, Trevor Killeen, Zeming Lin, Natalia Gimelshein, Luca Antiga, Alban
  Desmaison, Andreas Kopf, Edward Yang, Zachary DeVito, Martin Raison, Alykhan
  Tejani, Sasank Chilamkurthy, Benoit Steiner, Lu Fang, Junjie Bai, and Soumith
  Chintala.
\newblock Pytorch: An imperative style, high-performance deep learning library.
\newblock In H. Wallach, H. Larochelle, A. Beygelzimer, F. d\textquotesingle
  Alch\'{e}-Buc, E. Fox, and R. Garnett, editors, {\em Advances in Neural
  Information Processing Systems 32}, pages 8024--8035. Curran Associates,
  Inc., 2019.

\bibitem{tvloss}
Leonid~I. Rudin and S. Osher.
\newblock Total variation based image restoration with free local constraints.
\newblock {\em Proceedings of 1st International Conference on Image
  Processing}, 1:31--35 vol.1, 1994.

\bibitem{plenoxels}
{Sara Fridovich-Keil and Alex Yu}, Matthew Tancik, Qinhong Chen, Benjamin
  Recht, and Angjoo Kanazawa.
\newblock Plenoxels: Radiance fields without neural networks.
\newblock In {\em CVPR}, 2022.

\bibitem{colmap}
Johannes~L. Schönberger and Jan-Michael Frahm.
\newblock Structure-from-motion revisited.
\newblock In {\em CVPR}, pages 4104--4113, 2016.

\bibitem{lightfields}
Mohammed Suhail, Carlos Esteves, Leonid Sigal, and Ameesh Makadia.
\newblock Light field neural rendering.
\newblock Dec 2021.

\bibitem{refnerf}
Dor Verbin, Peter Hedman, Ben Mildenhall, Todd Zickler, Jonathan~T. Barron, and
  Pratul~P. Srinivasan.
\newblock {Ref-NeRF}: Structured view-dependent appearance for neural radiance
  fields.
\newblock {\em CVPR}, 2022.

\bibitem{ssim}
Zhou Wang, A.C. Bovik, H.R. Sheikh, and E.P. Simoncelli.
\newblock Image quality assessment: from error visibility to structural
  similarity.
\newblock {\em IEEE Transactions on Image Processing}, 13(4):600--612, 2004.

\bibitem{nerfmm}
Zirui Wang, Shangzhe Wu, Weidi Xie, Min Chen, and Victor~Adrian Prisacariu.
\newblock Ne{RF}$--$: Neural radiance fields without known camera parameters.
\newblock {\em arXiv preprint arXiv:2102.07064}, 2021.

\bibitem{scnerf}
Christopher Choy Animashree Anandkumar Minsu~Cho Yoonwoo~Jeong, Seokjun~Ahn and
  Jaesik Park.
\newblock Self-calibrating neural radiance fields.
\newblock In {\em ICCV}, 2021.

\bibitem{plenoctrees}
Alex Yu, Ruilong Li, Matthew Tancik, Hao Li, Ren Ng, and Angjoo Kanazawa.
\newblock {PlenOctrees} for real-time rendering of neural radiance fields.
\newblock In {\em ICCV}, 2021.

\bibitem{lpips}
Richard Zhang, Phillip Isola, Alexei Efros, Eli Shechtman, and Oliver Wang.
\newblock The unreasonable effectiveness of deep features as a perceptual
  metric.
\newblock 01 2018.

\bibitem{ate}
Zichao Zhang and Davide Scaramuzza.
\newblock A tutorial on quantitative trajectory evaluation for
  visual(-inertial) odometry.
\newblock In {\em 2018 IEEE/RSJ International Conference on Intelligent Robots
  and Systems (IROS)}, pages 7244--7251, 2018.

\end{thebibliography}
}
\end{document}